\documentclass[11pt,a4paper]{article}
\usepackage[hyperref]{acl2020}
\usepackage[whole]{bxcjkjatype}
\usepackage{times}
\usepackage{latexsym}

\usepackage{microtype}

\aclfinalcopy %

\usepackage{amsmath,amssymb,mathtools,bm}

\newcommand{\argmax}{\mathop{\rm arg~max}\limits}

\usepackage{enumitem}

\usepackage{graphicx}
\usepackage{booktabs,multirow}
\usepackage{caption}
\usepackage[margin=5pt]{subcaption}

\usepackage{ulem}
\newcommand{\tablabel}[1]{\footnotesize\texttt{#1}}

\newcommand*{\ka}[1]{\textcolor{red}{\bf #1}}

\title{Embeddings of Label Components for Sequence Labeling:\\A Case Study of Fine-grained Named Entity Recognition}

\author{Takuma\,Kato$^{1}$\hspace{1em}
        Kaori\,Abe$^{1,2}$\hspace{1em}
        Hiroki\,Ouchi$^{2,1}$\hspace{1em}
        Shumpei\,Miyawaki$^{1}$\hspace{1em} \\
        \textbf{Jun\,Suzuki$^{1,2}$}\hspace{1em}
        \textbf{Kentaro\,Inui$^{1,2}$} \\
  {$^1$ Tohoku University} \hspace{1.0cm} {$^2$ RIKEN}\\
  \normalsize{\texttt{\{takuma.kato,abe-k,miyawaki.shumpei,jun.suzuki,inui\}@ecei.tohoku.ac.jp}} \\
  \normalsize{\texttt{hiroki.ouchi@riken.jp}}}

\date{}

\begin{document}
\maketitle
\begin{abstract}
In general, the labels used in sequence labeling consist of different types of elements.
For example, IOB-format entity labels, such as \texttt{B-Person} and \texttt{I-Person}, can be decomposed into span (\texttt{B} and \texttt{I}) and type information (\texttt{Person}).
However, while most sequence labeling models do not consider such label components, the shared components across labels, such as \texttt{Person}, can be beneficial for label prediction.
In this work, we propose to integrate label component information as embeddings into models.
Through experiments on English and Japanese fine-grained named entity recognition, we demonstrate that the proposed method improves performance, especially for instances with low-frequency labels.
\end{abstract}

\section{Introduction}\label{sec:introduction}
Sequence labeling is a problem in which a label is assigned to each word in an input sentence. 
In many label sets, each label consists of different types of elements.
For example, IOB-format entity labels \cite{ramshaw-marcus-1995-text}, such as \texttt{B-Person} and \texttt{I-Location}, can be decomposed into span (e.g., \texttt{B}, \texttt{I} and \texttt{O}) and type information (e.g., \texttt{Person} and \texttt{Location}).
Also, morphological feature tags \cite{more-etal-2018-conll}, such as \texttt{Gender=Masc|Number=Sing}, can be decomposed into gender, number and other information.

General sequence labeling models \cite{ma-hovy-2016-end,lample-etal-2016-neural,chiu-nichols-2016-named}, however, do not consider such components.
Specifically, the probability that each word is assigned a label is computed on the basis of the inner product between word representation and label embedding (see Equation~\ref{eq:prob} in Section~\ref{sec:base}).
Here, the label embedding is associated with each label and independently trained without considering its components.
This means that labels are treated as mutually exclusive.
In fact, labels often share some components.
Consider the labels \texttt{B-Person} and \texttt{I-Person}.
They share the component \texttt{Person}, and injecting such component information into the label embeddings can improve the generalization performance.

Motivated by this, we propose a method that shares and learns the embeddings of label components (see details in Section~\ref{sec:proposed}).
Specifically, we first decompose each label into its components.
We then assign an embedding to each component and summarize the embeddings of all the components into one as a label embedding used in a model.
This component-level operation enables the model to share information on the common components across label embeddings.

To investigate the effectiveness of our method, we take the task of fine-grained Named Entity Recognition (NER) as a case study.
Typically, in this task, a large number of entity-type labels are predefined in a hierarchical structure, and intermediate type labels can be used as label components, as well as leaf type labels and B/I-labels. 
In this sense, the fine-grained NER can be seen as a good example of the potential applications of the proposed method. 
Furthermore, some entity labels occur more frequently than others.
An interesting question is whether our method of label component sharing exhibits an improvement in recognizing entities of infrequent labels. 
In our experiments, we use the English and Japanese NER corpora with the Extended Named Entity Hierarchy~\cite{DBLP:conf/lrec/SekineSN02} including 200 entity tags.
To sum up, our main contributions are as follows: (i) we propose a method that shares and learns label component embeddings, and (ii) through experiments on English and Japanese fine-grained NER, we demonstrate that the proposed method achieves better performance than a standard sequence labeling model, especially for instances with low-frequency labels.

\begin{figure}
  \centering
  \includegraphics[width=7cm
  ]{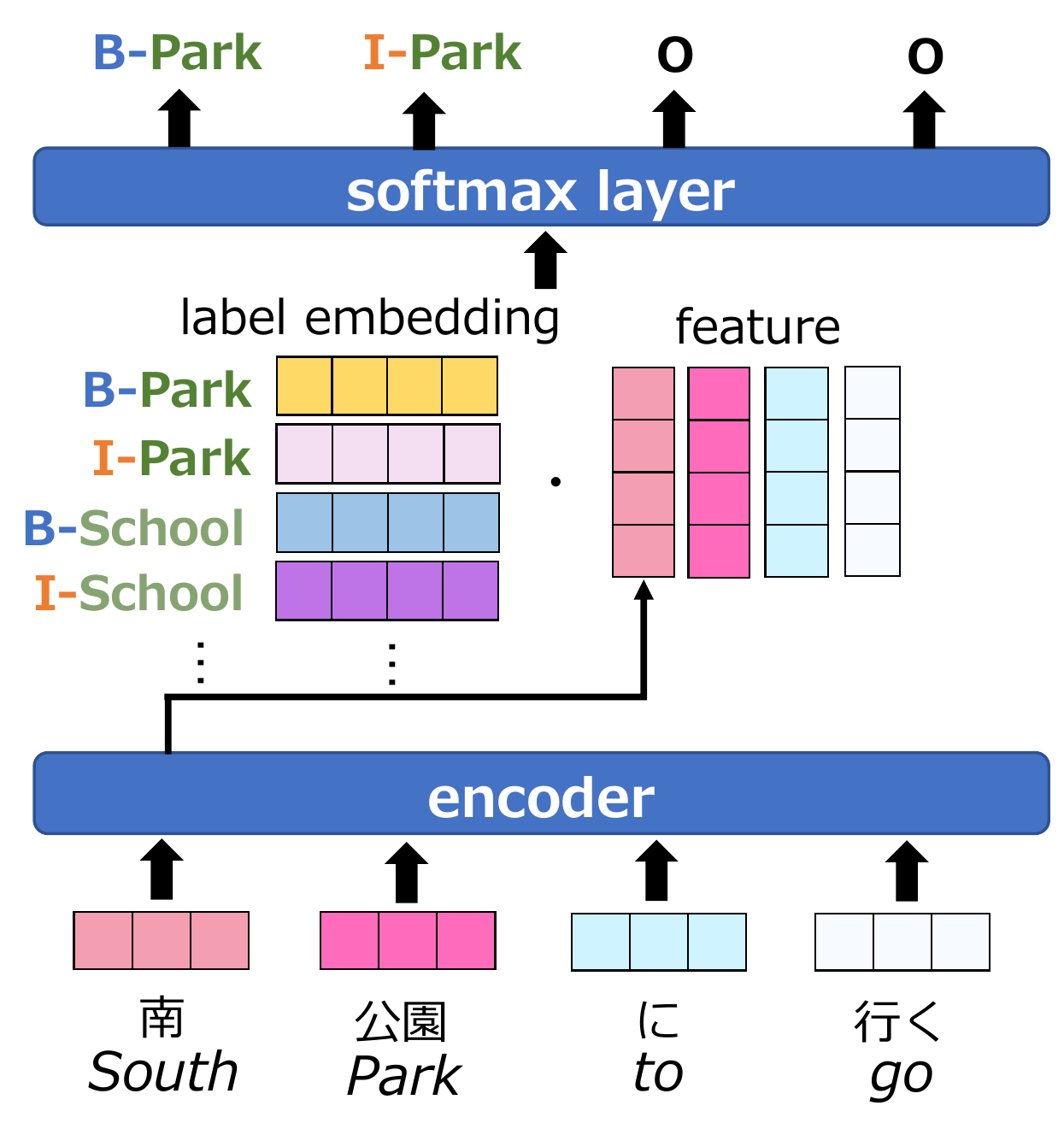}
\caption{Overview of a standard sequence labelling model. Each label (e.g., B-Park) is annotated as a single unit, disregarding its inner structure (``B" and ``Park").}
\label{fig:NER_example}
\end{figure}
\section{Methods}\label{sec:method}

\subsection{Baseline model}\label{sec:base}
We describe our baseline model in Figure~\ref{fig:NER_example}.
Given an input sentence, the encoder converts each word into its feature vector.
Then, the inner product between each feature vector and label embedding is calculated for computing the label distribution.
Finally, the IOB2-format label \cite{ramshaw-marcus-1995-text} with the highest probability is assigned to each token.
The label \texttt{B-Park}, indicating the leftmost token of some entity, is assigned to 南 (\textit{South}), and \texttt{I-Park}, indicating the token inside some entity, is assigned to 公園 (\textit{Park}).
The label \texttt{O}, indicating the token outside entities, is assigned to に (\textit{to}) and 行く (\textit{go}).

Formally, for each word $x_i$ in the input sentence $X=(x_1,x_2,\dots,x_n)$, the model outputs the label $\hat{y}_i$ with the highest probability:
\begin{eqnarray}
\label{eq:predict}
    \hat{y_i} = \argmax_{y\in\mathcal{Y}}P(y|x_i,X),
\end{eqnarray}

\noindent
where $\mathcal{Y}$ is a label set defined in each data set.
The probability distribution is calculated as
\begin{eqnarray}
\label{eq:prob}
    P(y|x_i,X) = \frac{\exp(\mathbf{W}[y] \cdot \mathbf{f}(x_i,X))}{\displaystyle \sum_{y' \in \mathcal{Y}}\exp(\mathbf{W}[y'] \cdot \mathbf{f}(x_i,X))},
\end{eqnarray}

\noindent
where $\mathbf{W} \in \mathbb{R}^{|\mathcal{Y}| \times D}$ is a weight matrix for the label set $\mathcal{Y}$.\footnote{$D$ is the number of dimensions of each weight vector.}
Each row of this matrix is associated with each label $ y \in \mathcal {Y} $, and $ \mathbf{W}[y]$ represents the $ y $-th row vector.
$\mathbf{f}(x,X)$ represents the vector encoded by a neural-network-based encoder.
\begin{figure}[!t]
  \centering
  \includegraphics[width=\hsize]{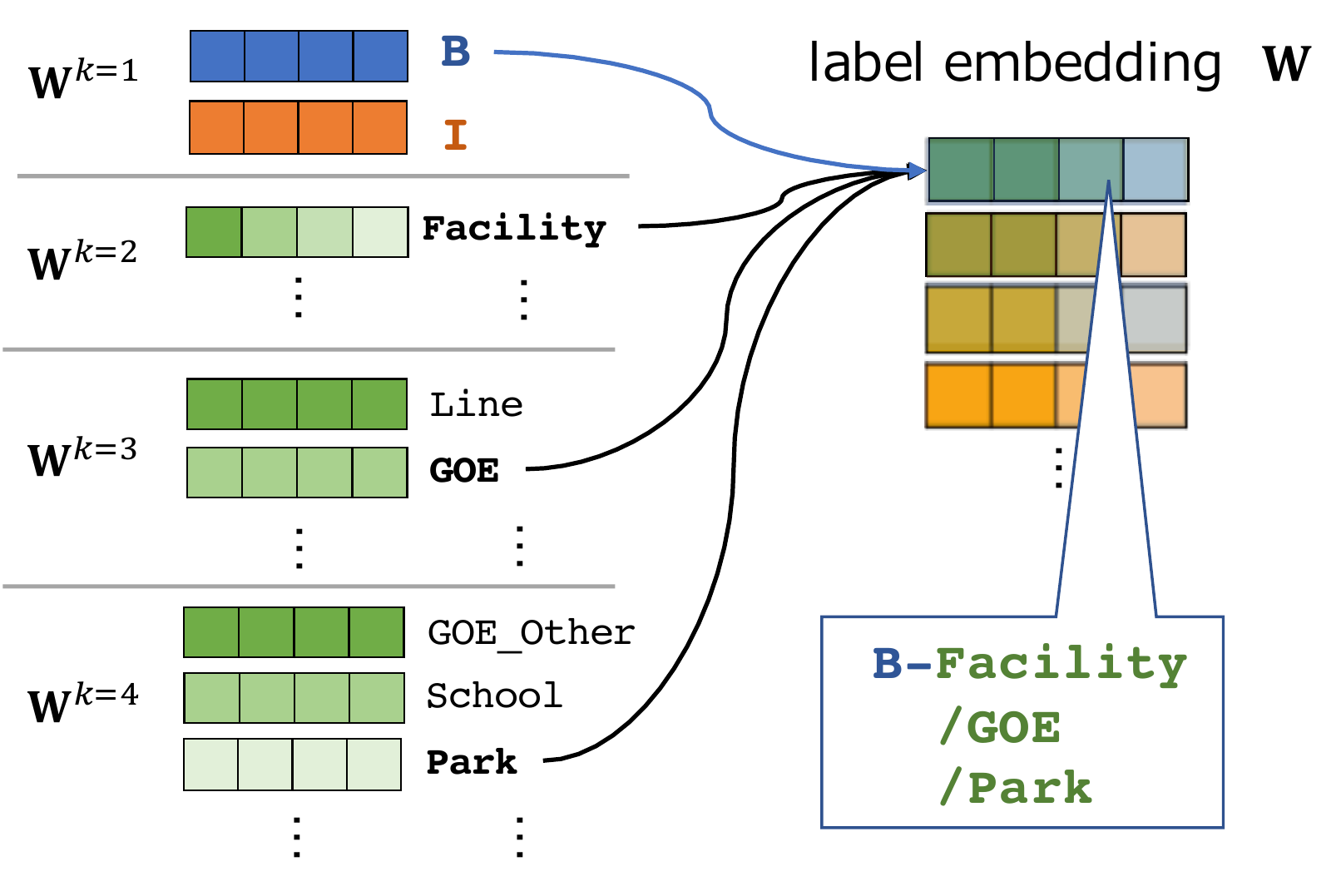}
\caption{Label embedding calculation. Each label embedding is calculated from its component embeddings.}
  \label{fig:proposed_method}
\end{figure}

\subsection{Embeddings of label components}
\label{sec:proposed}

We propose to integrate label component information as embeddings into models. This procedure consists of two steps: (i) \textit{label decomposition} and (ii) \textit{label embedding calculation}.

\paragraph{Label decomposition}
We first decompose each label into its components.
Each label consists of multiple types of components. Consider the following example.
\begin{align}
\nonumber \texttt{B-Park} = \{ \texttt{B}, \texttt{Park} \}
\end{align}

\noindent
The labels defined in a general entity tag set consist of the IOB (e.g.,~\texttt{B}) and entity (e.g.,~\texttt{Park}) component types.
Consider another example.
\begin{align}
\nonumber & \texttt{B-Facility/GOE/Park} = \\
\nonumber & \{ \texttt{B}, \texttt{Facility}, \texttt{GOE}, \texttt{Park} \}
\end{align}

\noindent
The labels defined in the Extended Named Entity tag set \cite{DBLP:conf/lrec/SekineSN02} consist of the four component types:  IOB (e.g.,~\texttt{B}), top layer of the entity tag hierarchy (e.g.,~\texttt{Facility}), second layer (e.g.,~\texttt{GOE}) the third layer (e.g.,~\texttt{Park}).
In this way, we can regard each label as a set of components (type–value pairs).

Formally, $K$ components of each label $y$ will be denoted by $\mathcal{C}^y = \{ c_k \}^K_{k=1}$, where $c_k$ is the index associated with the value of each component type $k$.
The above two examples are represented as  $\mathcal{C}^{y=\texttt{B-Park}} = \{c_1=\texttt{B}, c_2=\texttt{Park} \}$ and $\mathcal{C}^{y=\texttt{B-Facility/GOE/Park}} = \{c_1=\texttt{B}, c_2=\texttt{Facility}, c_3=\texttt{GOE}, c_4=\texttt{Park} \}$.
This formalization is applicable to arbitrary label sets whose label consists of type-value components.

\paragraph{Label embedding calculation}
We then assign an embedding (i.e., trainable vector representation) to each label component and %
combining the embeddings of all the components within a label into one label embedding.
In this study, we investigate two types of typical summarizing techniques: (a) summation and (b) concatenation.
\paragraph{(a) Summation}
The embedding of each label, $\mathbf{W}[y]$, is calculated by summing the embeddings of its components:
\begin{equation}
\label{eq:proposed-sum}
    \mathbf{W}[y] =  \sum_{c_k \in \mathcal{C}^y} \mathbf{W}^k[c_k].
\end{equation}

\noindent
Here, $\mathbf{W}^k$ is an embedding matrix for each component type $k$, and $\mathbf{W}^k[c_k]$ denotes the $c_k$-th row vector.
Figure~\ref{fig:proposed_method} illustrates this calculation process.
The label \texttt{B-Facility/GOE/Park} consists of four components (i.e., \texttt{B}, \texttt{Facility}, \texttt{GOE} and \texttt{Park}), each $c_k$ value of which is associated with a row vector of each matrix $\mathbf{W}^k$.

\paragraph{(b) Concatenation}
The embedding of each label, $\mathbf{W}[y]$, is calculated by concatenating the embeddings of its components:
\begin{equation}
\label{eq:proposed-concat}
    \mathbf{W}[y] = [\mathbf{W}^1[c_1], \dots, \mathbf{W}^K[c_K]].
\end{equation}

\noindent
Here, similarly to $\mathbf{W}^k$ is an embedding matrix for each component type $k$ Equation~\ref{eq:proposed-sum}.
Unlike Equation~\ref{eq:proposed-sum}, the label component embeddings are concatenated into one embedding.
Compared with the summation, one disadvantage of the concatenation is memory efficiency: the number of dimensions of the label embeddings increases according to the number of label components $K$.\\

Our label embedding calculation enables models to share the embeddings of label components commonly shared across labels.
For example, the embeddings of both \texttt{B-Facility/GOE/Park} and \texttt{B-Facility/GOE/School} are calculated by
adding the embeddings of the shared components (i.e., \texttt{B}, \texttt{Facility} and \texttt{GOE}).
Equations~\ref{eq:proposed-sum}~and~\ref{eq:proposed-concat} can be regarded as a general form of the hierarchical label matrix proposed by \citet{DBLP:conf/eacl/InuiRSS17} because our method can treat not only hierarchical structures but also any type of type–value set, such as morphological feature labels (e.g. \texttt{Gender=Masc|Number=Sing}).

\section{Experiments}

\subsection{Settings}

\paragraph{Dataset}

\begin{table*}[!t]
\centering
\begin{tabular}{lrrrr}
\toprule
\multicolumn{1}{c}{\multirow{2}{*}{Dataset}} & \multicolumn{2}{c}{English}                                              & \multicolumn{2}{c}{Japanese}                                             \\
\cmidrule{2-5}
\multicolumn{1}{c}{}                         & \multicolumn{1}{c}{\# of Sentences} & \multicolumn{1}{c}{\# of Entities} & \multicolumn{1}{l}{\# of Sentences} & \multicolumn{1}{l}{\# of Entities} \\
\midrule
train                                        & 14176                               & 27686                              & 34784                                 & 72318                                \\
dev                                          & 1573                                & 3032                               & 7009                                & 11954                               \\
test                                         & 3942                                & 7682                               & 6783                                & 11669 \\
\bottomrule
\end{tabular}
\caption{Statistics of the datasets.}
\label{tab:extended_NER_corpus}
\end{table*}
\begin{table}[!t]
\centering
\scalebox{0.82}{
\begin{tabular}{llrrrr}
\toprule
\multicolumn{2}{l}{Frequency Classes} &\multicolumn{2}{c}{English} & \multicolumn{2}{c}{Japanese}\\
 \cmidrule{3-6}
& & Dev & Test & Dev & Test\\
\midrule
Low &（0$\sim$100）  & 1125&  2798 & 666 & 619\\
Middle &（101$\sim$500）&1224 & 3128  & 2,875    & 2,531  \\
High &（501$\sim$）  & 683 & 1756 & 8,413    & 8,519\\ \bottomrule
\end{tabular}
}
\caption{Details of frequency classes.}
\label{tab:label_frequency_detail}
\end{table}

We use the Extended Named Entity Corpus for English\footnote{We e-mailed the authors of \cite{mai-etal-2018-empirical} and received the English dataset.} and Japanese.\footnote{\url{https://www.gsk.or.jp/catalog/gsk2014-a/}} fine-grained NER~\cite{mai-etal-2018-empirical}
In this dataset, each NE is assigned one of 200 entity labels defined in the Extended Named Entity Hierarchy~\cite{DBLP:conf/lrec/SekineSN02}.
For the English dataset, we follow the training/development/test split defined by \citet{mai-etal-2018-empirical}.
For the Japanese dataset, we follow the training/development/test split of Universal Dependencies (UD) Japanese-BCCWJ.~\cite{asahara-etal-2018-universal}\footnote{\url{https://github.com/UniversalDependencies/UD_Japanese-BCCWJ}}
Table~\ref{tab:extended_NER_corpus} shows the statistics of the dataset.

\paragraph{Data statistics}
There is a gap between the frequencies, i.e., how many times each label appears in the training set.
We categorize each label into three classes on the basis of its frequency, shown in Table~\ref{tab:label_frequency_detail}.
For example, if a label appears $0$–$100$ times in the training set, it is categorized into the ``Low" class.
Moreover, we denote how many times entities with the labels belonging to each frequency class appear in the development or test set.
To better understand the model behavior, we investigate the performance of each frequency class.

\paragraph{Model setup}
As the encoder $\mathbf{f}(x, X)$ in Equation~\ref{eq:prob} in Section~\ref{sec:base}, we use BERT\footnote{We use the open-source NER model utilizing BERT: \url{https://github.com/kamalkraj/BERT-NER}.}~\cite{DBLP:conf/naacl/DevlinCLT19}, which is a state-of-the-art language model.\footnote{The state of the art model on the Extended Named Entity Corpus is the LSTM + CNN + CRF model that uses dictionary information~\cite{mai-etal-2018-empirical}}
As the baseline model, we use the general label embedding matrix without considering label components, i.e., each label embedding $\mathbf{W}[y]$ in Equation~\ref{eq:prob} is randomly initialized and independently learned.
In contrast, our proposed model calculates the label embedding matrix from label components (Equations~\ref{eq:proposed-sum}~and~\ref{eq:proposed-concat}).
The only difference between these models is the label embedding matrix, so if a performance gap between them is observed, it stems from this point.

\paragraph{Hyperparameters}
The overall settings of hyperparameters are the same between the baseline and the proposed model.
For English, we use the BERT pre-trained on BooksCorpus and English Wikipedia~\cite{DBLP:conf/naacl/DevlinCLT19}.
For Japanese, we use the BERT pre-trained on Japanese Wikipedia~\cite{Shibata-etal-2019-bert}. We fine-tune them on the Extended NER corpus for solving fine-grained NER.
We set the training epochs to $20$ in fine-tuning. 
Both the baseline and the proposed models are trained to minimize cross-entropy loss during training.
We set a batch size of $32$ and a learning rate of $5.0 \times 10^{-5}$ using Adam~\cite{DBLP:journals/corr/KingmaB14} for the optimizer.
We choose the dropout rate from among $\{0.1, 0.3, 0.5\}$ on the basis of the F$_1$ scores in each development set.\footnote{In our experiments, we found that the models trained with the dropout rate of $0.1$ achieved the best performance on each development set.}
We set the number of dimensions of the hidden states in BERT.
In the baseline model, we set the number of dimensions of the label embedding $\mathbf{W}$ in Equation~\ref{eq:prob} to $768$.
In the proposed models, we also use the same dimension size $768$ for $\mathbf{W}$ in Equations~\ref{eq:proposed-sum}~and~\ref{eq:proposed-concat}.

\subsection{Results}
\label{sec:result}

We report averaged F$_1$ scores across five different runs of the model training with random seeds.
Table~\ref{tab:f1_per_frequency_result} shows F$_1$ scores for overall classes and each label frequency class on each test set.

\paragraph{Overall performance}
For the overall labels, the proposed models (\textsc{Proposed:Sum} and \textsc{Proposed:Concat}) outperformed the baseline model on English and Japanese datasets.
These results suggest the effectiveness of our proposed method for calculating the label embeddings from label components.

\begin{table*}[!t]
\centering
\begin{tabular}{lcccc}
\toprule
           & Low & Middle & High & Overall \\
\midrule
\multicolumn{5}{c}{English} \\
\midrule
\textsc{Baseline}    & 79.83±0.27 & 80.29±0.46              & 90.82±0.32 & 84.99±0.27                          \\
\textsc{Proposed:Sum}    & \textbf{81.15}±0.24             & \textbf{80.99}±0.27              & \textbf{90.87}±0.26  & \textbf{85.67}±0.13                          \\
\textsc{Proposed:Concat} & 80.40±0.38               & 80.31±0.28              & 90.75±0.23 & 85.20±0.16                          \\
\midrule
\multicolumn{5}{c}{Japanese} \\
\midrule
\textsc{Baseline}    & 44.39±0.29 & 51.73±0.50              & 70.82±0.32 & 68.06±0.27                          \\
\textsc{Proposed:Sum}    & \textbf{45.34}±0.91               & \textbf{51.93}±0.66              & \textbf{71.04}±0.49  & \textbf{68.34}±0.41                          \\
\textsc{Proposed:Concat} & 44.76±1.12               & 51.45±0.40              & 70.52±0.29 & 67.77±0.23                          \\
\bottomrule
\end{tabular}
\caption{Comparison between the baseline and proposed models. Cells show the F$_1$ scores and standard deviations on each test set.}
\label{tab:f1_per_frequency_result}
\end{table*}

\paragraph{Performance for each frequency class}
For all the label frequency classes, the proposed model with summation (\textsc{Proposed:Sum}) yielded the best results among the three models. 
In particular, for low-frequency labels, the proposed model with summation (\textsc{Proposed:Sum}) achieved a remarkable improvement of F$_1$ compared with the baseline model.
Also, the proposed model with concatenation (\textsc{Proposed:Concat}) achieved an improvement of F$_1$.
These results suggest that exploiting label embeddings of the components shared across labels improves the generalization performance,  especially for low-frequency labels.

\subsection{Analysis}

\begin{table}[!t]
\centering
\begin{tabular}{lrr}
\toprule
                & \multicolumn{1}{c}{English} & \multicolumn{1}{c}{Japanese} \\
\midrule
Baseline        & 76.58±0.26                  & 49.66±0.68                   \\
Proposed:Sum    & \textbf{77.76}±0.30                  & \textbf{50.05}±1.19                   \\
Proposed:Concat & 76.77±0.71                  & 49.31±1.12                   \\
\bottomrule
\end{tabular}
\caption{Comparison between the baseline and the proposed models in the Low frequency class.}
\label{tab:low-frequency label result}
\end{table}

Recall that the entity tag set used in the datasets has a hierarchical structure.
This means that label components at higher layers appear more frequently than those at lower layers and are shared across many labels.
As shown in Table~\ref{tab:f1_per_frequency_result}, the proposed models achieve performance improvements for low-frequency labels.
Here, we can expect that the embeddings of high-frequency shared label components help the model correctly predict the low-frequency labels.
To verify this hypothesis, we compare between F$_1$ scores of the baseline and proposed models, shown in Table~\ref{tab:low-frequency label result}.
Here, the targets to investigate are the three-layered, low-frequency labels\footnote{We exclude the labels that consist of only two layers, such as \texttt{Timex/Date}.} that have a high-frequency, second layer component.\footnote{In this paper, we also regard the second-layer components appearing over 100 times in the training set as high-frequency.}
As shown in Table~\ref{tab:low-frequency label result}, the \textsc{Proposed:Sum} model outperformed the baseline model. 
This indicates that for predicting low-frequency labels, it is effective for the model to use shared components.
On the other hand, the \textsc{Proposed:Concat} model underperformed the baseline model. One possible reason is that the model obtains less information by concatenating label embeddings than by summing them.

\subsection{Visualization of label embedding spaces}
\begin{figure*}[!t]

\begin{tabular}{cc}

\begin{minipage}[t]{0.48\hsize}
\centering
\includegraphics[width=\hsize]{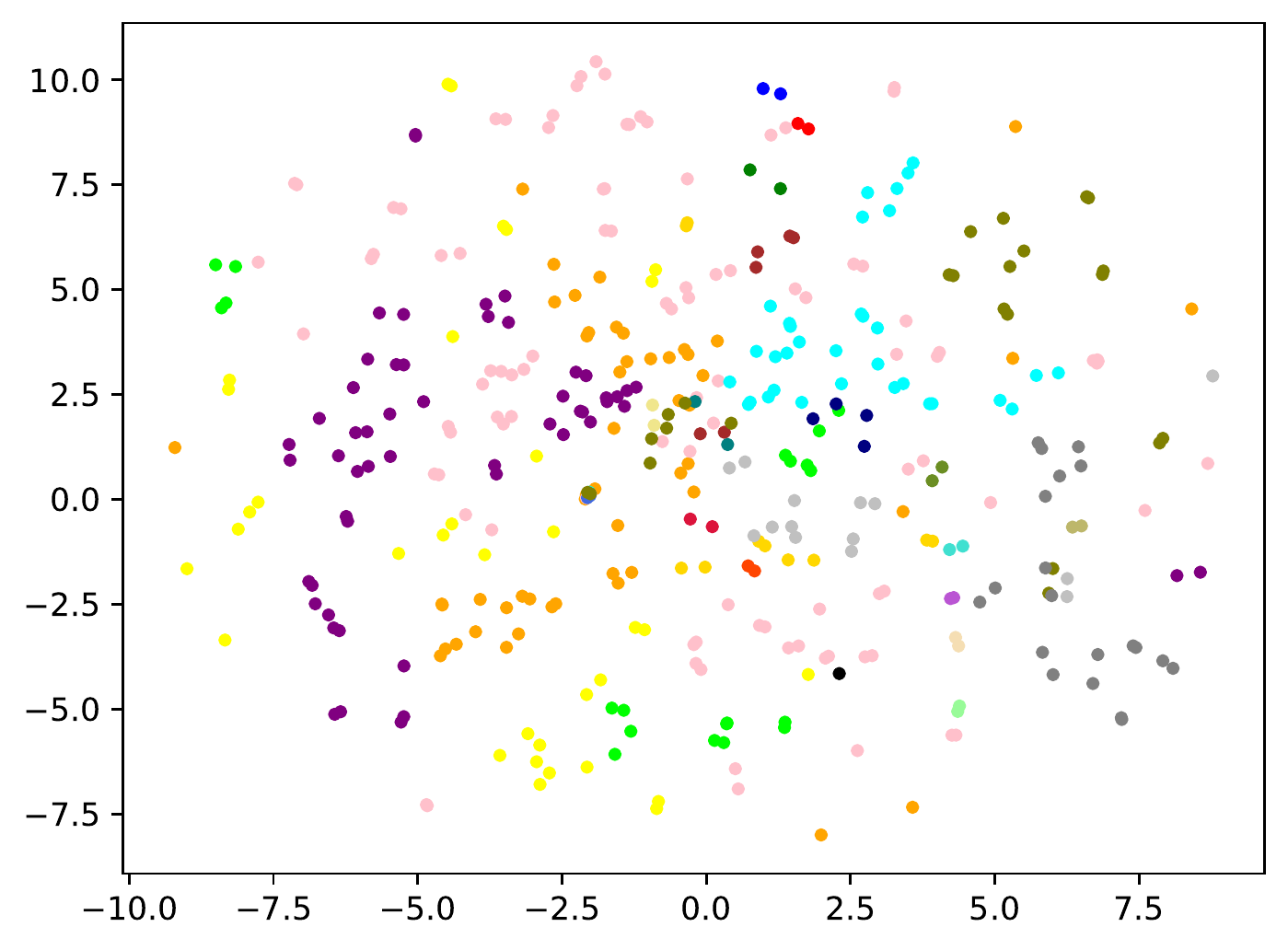}
\vspace{-0.7cm}
\subcaption{\textsc{Baseline}}
\label{fig:label_embedding_space_a}
\end{minipage}
&
\begin{minipage}[t]{0.48\hsize}
\centering
\includegraphics[width=\hsize]{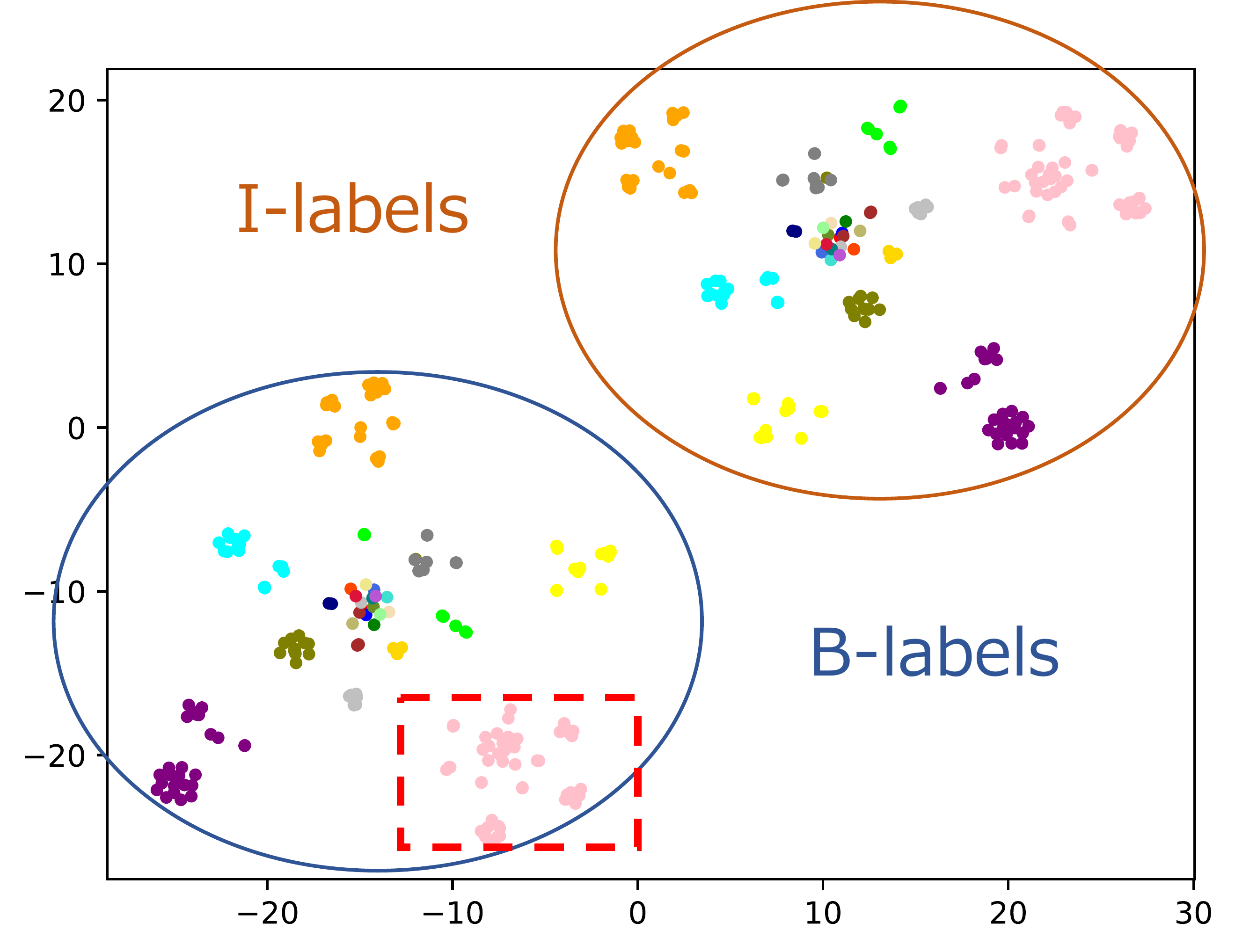}
\vspace{-0.7cm}
\subcaption{\textsc{Proposed:Sum}}
\label{fig:label_embedding_space_b}
\end{minipage}
\\
\multicolumn{2}{c}
{
\begin{minipage}[t]{\hsize}
\centering
  \includegraphics[width=10cm]{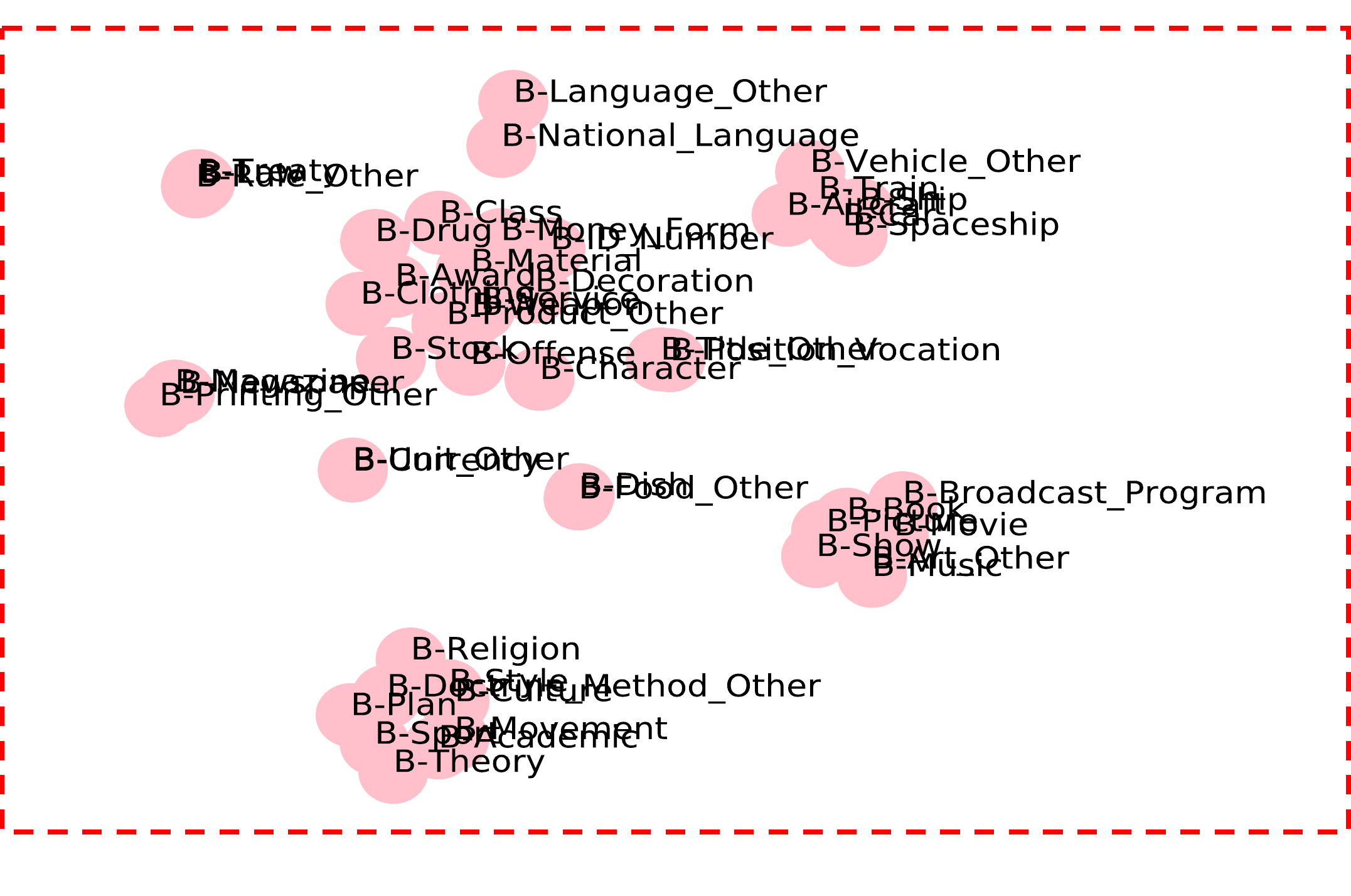}
\vspace{-0.5cm}
\subcaption{Enlarged view of a cluster in (b). The embeddings of the labels sharing the top layer label \texttt{Product} form this cluster.}
\label{fig:label_embedding_space_c}
\end{minipage}
}
\end{tabular}
\caption{Visualization of the label embedding space. The same color represents the labels that have the same hierarchical top layer label.
}
\label{fig:label_embedding_space}
\end{figure*}

To better understand the label embeddings created from the label components by our proposed method, we visualize the learned label embeddings.
Specifically, we hypothesize that the embeddings of the labels sharing label components are close to each other and form clusters in the embedding space if they successfully encode the shared label component information.
To verify this hypothesis, we 
use the t-SNE algorithm~\cite{vanDerMaaten2008} to map the label embeddings learned by the baseline and proposed models onto the two-dimensional space, shown in Figure~\ref{fig:label_embedding_space}.
As we expected, some clusters were formed in the label embedding space learned by the proposed model, shown in Figure~\ref{fig:label_embedding_space_b}, while there is no distinct cluster in the one learned by the baseline, shown in Figure~\ref{fig:label_embedding_space_a}.
By looking at them in detail, we obtained two findings.
First, in the embedding space learned by the proposed model, we found that two distinct clusters were formed corresponding to the two span labels (i.e. \texttt{B} and \texttt{I}).
Second, the labels that have the same top layer label (represented in the same color) also formed some smaller clusters within the \texttt{B} and \texttt{I}-label clusters.
For example, Figure~\ref{fig:label_embedding_space_c} shows the \texttt{Product} cluster whose members are the labels sharing the top layer label \texttt{Product}.
From these figures, we could confirm that the embeddings of the labels sharing label components (span and upper-layer type labels) form the clusters.

\section{Related work}

Sequence labeling has been widely studied and applied to many tasks, such as Chunking \cite{ramshaw-marcus-1995-text,hashimoto-etal-2017-joint}, NER \cite{ma-hovy-2016-end,chiu-nichols-2016-named} and Semantic Role Labeling (SRL) \cite{zhou-xu-2015-end,he-etal-2017-deep}. 
In English fine-grained entity recognition, \citet{DBLP:conf/aaai/LingW12} created a standard fine-grained entity typing dataset with multi-class, multi-label annotations.
\citet{DBLP:conf/acl/RinglandDHKPC19} developed a dataset for nested NER dataset.
These datasets independently handle each label without considering label components.
In Japanese NER, \citet{misawa-etal-2017-character} combined word and character information to improve performance.
\citet{mai-etal-2018-empirical} reported that dictionary information improves the performance of fine-grained NER.
Their methods do not consider label components and are orthogonal to our method.

Some existing studies take shared components (or information) across labels into account.
In Entity Typing, \citet{DBLP:conf/coling/MaCG16} and \citet{DBLP:conf/eacl/InuiRSS17} proposed to calculate entity label embeddings by considering a label hierarchical structure. 
While their method is limited to only a hierarchical structure, our method can be applied to any set of components and can be regarded as a general form of their method.
In multi-label classification, \citet{DBLP:conf/icdm/ZhongXDZ18} assumed that the labels co-occurring in many instances are correlated with each other and share some common features, and proposed a method that learns a feature (label embedding) space where such co-occurring labels are close to each other.
The work of \citet{matsubayashi-etal-2009-comparative} is the closest to ours in terms of decomposing the features of labels.
They regard an original label comprising a mixture of components as a set of multiple labels and made models that are able to exploit the multiple components to effectively learn in the SRL task.

\section{Conclusion}
We proposed a method that shares and learns the embeddings of label components.
Through experiments on English and Japanese fine-grained NER, we demonstrated that our proposed method improves the performance, especially for instances with low-frequency labels.
For future work, we envision to apply our method to other tasks and datasets and investigate the effectiveness.
Also, we plan to extend the simple label embedding calculation methods to more sophisticated ones.

\section*{Acknowledgments}
This work was partially supported by JSPS KAKENHI Grant Number JP19H04162 and JP19K20351. This work was also partially supported by a Bilateral Joint Research Program between RIKEN AIP Center and Tohoku University.
We would like to thank the members of Tohoku NLP Laboratory, the anonymous reviewers, and the SRW mentor Gabriel Stanovsky for their insightful comments.
We also appreciate Alt inc. for providing the corpus of English extended named entity data.

\bibliography{references}
\bibliographystyle{acl_natbib}

\clearpage
\appendix

\section{Appendices}
\subsection{Additional results}
\begin{table}[h]
\centering
\scalebox{0.72}{
\begin{tabular}{lrrr}
\toprule
                & \multicolumn{1}{c}{Top}                          & \multicolumn{1}{c}{Second}                         & \multicolumn{1}{c}{Third}                          \\
\midrule
                \multicolumn{4}{c}{English}                                                                      \\
\midrule
\textsc{Baseline}        & \multicolumn{1}{r}{90.01±0.27} & \multicolumn{1}{r}{86.69±0.32} & \multicolumn{1}{r}{83.22±0.28} \\
\textsc{Proposed:Sum}    & \multicolumn{1}{r}{\textbf{90.53}±0.06} & \multicolumn{1}{r}{\textbf{87.53}±0.11} & \multicolumn{1}{r}{\textbf{83.87}±0.20} \\
\textsc{Proposed:Concat} & \multicolumn{1}{r}{90.28±0.09} & \multicolumn{1}{r}{87.04±0.13} & \multicolumn{1}{r}{83.18±0.30} \\
\midrule
                \multicolumn{4}{c}{Japanese}                                                                     \\
\midrule
\textsc{Baseline}        & 72.68±0.20                     & 66.22±0.36                     & 66.84±0.34                     \\
\textsc{Proposed:Sum}    & \textbf{73.13}±0.43                     & \textbf{66.37}±0.42                     & \textbf{67.00}±0.59                     \\
\textsc{Proposed:Concat} & 72.50±0.30                     & 66.19±0.24                     & 66.42±0.49 \\
\bottomrule
\end{tabular}
}
\caption{Comparison between the baseline and proposed models for the labels at each hierarchical layer.}
\label{tab:hier_result}
\end{table}
\begin{table}[h]
\centering
\scalebox{0.9}{
\begin{tabular}{lrr}
\toprule
                & \multicolumn{1}{c}{English} & \multicolumn{1}{c}{Japanese} \\
\midrule
\textsc{Baseline}        & 96.32±0.10                  & 84.74±0.18                   \\
\textsc{Proposed:Sum}    & 96.31±0.11                  & 85.01±0.15                   \\
\textsc{Proposed:Concat} & 96.27±0.07                  & 84.83±0.11                   \\
\bottomrule
\end{tabular}
}
\caption{Comparison between the baseline and the proposed models in span (only considering B, I labels).}
\label{tab:span_result}
\end{table}
\paragraph{Performance for each hierarchical category}
Table~\ref{tab:hier_result} shows F$_1$ scores for each hierarchical category.
The proposed model with summation (\textsc{Proposed:Sum}) outperformed the other models in all the hierarchical categories. %
For the labels at the top layer, in particular, \textsc{Proposed:Sum} achieved an improvement of the F$_1$ scores by a large margin on the Japanese dataset.

\paragraph{Performance for entity span boundary match}
Table~\ref{tab:span_result} shows F$_1$ scores for entity span boundary match, where we regard a predicted boundary (i.e., B and I) as correct if it matches the gold annotation regardless of its entity type label.
The performance of the proposed models was comparable to the baseline model.
This indicates that there is a performance difference not in identification of entity spans (entity detection) but in identification of entity types (entity typing).
\subsection{Case study}
\begin{table}[ht]
\centering
\scalebox{0.57}{
\begin{tabular}{lllll}
\toprule
\textbf{Example (a)} & \multicolumn{4}{l}{\uuline{下呂}\ \uuline{温泉}\ 発祥\ の\ 地・・・} \\
& \multicolumn{4}{l}{(The birthplace of \uuline{Gero Spa} ... )} \\
\midrule
\textsc{Entity}  & \multicolumn{2}{l}{下呂 (\textit{Gero})} & \multicolumn{2}{l}{温泉 (\textit{Spa})}\\
\textsc{Gold} & \multicolumn{2}{l}{\tablabel{\underline{B-Location/Spa}}} & \multicolumn{2}{l}{\tablabel{\underline{I-Location/Spa}}}\\
\textsc{Baseline} & \multicolumn{2}{l}{\tablabel{\underline{B-}Facility/Facility\_Other}} & \multicolumn{2}{l}{\tablabel{\underline{I-}Facility/Facility\_Other}}\\
\textsc{Proposed:Sum} & \multicolumn{2}{l}{\tablabel{\underline{B-Location/Spa}}} & \multicolumn{2}{l}{\tablabel{\underline{I-Location/Spa}}} \\
\midrule
\textbf{Example (b)} & \multicolumn{4}{l}{・・・ where\ \uuline{clavaviridae}\ derives\ from\ .} \\
\midrule
\textsc{Entity}  & \multicolumn{4}{l}{clavaviridae}\\
\textsc{Gold} & \multicolumn{4}{l}{\tablabel{\underline{B-Natural\_Object/Living\_Thing/Living\_Thing\_Other}}}\\
\textsc{Baseline} & \multicolumn{4}{l}{\tablabel{\underline{B-}Location/Astral\_Body/Constellation}}\\
\textsc{Proposed:Sum} & \multicolumn{4}{l}{\tablabel{\underline{B-Natural\_Object/Living\_Thing/Living\_Thing\_Other}}}\\
\midrule
\textbf{Example (c)} & \multicolumn{4}{l}{・・・\uuline{あお白い}\ 日\ の\ 光\ ・・・} \\
& \multicolumn{4}{l}{(... the pale sunlight ... )} \\
\midrule
\textsc{Entity}  & \multicolumn{4}{l}{あお白い (\textit{pale})} \\
\textsc{Gold} & \multicolumn{4}{l}{\tablabel{\underline{B-Color/Color\_Other}}} \\
\textsc{Baseline} & \multicolumn{4}{l}{\tablabel{O}} \\
\textsc{Proposed:Sum} & \multicolumn{4}{l}{\tablabel{\underline{B-Color}/Nature\_Color}} \\
\bottomrule
\end{tabular}
}
\caption{Examples of both model outputs in fine-grained NER.}
\label{tab:result_example}
\end{table}
We observe actual examples predicted by the proposed model with summation, shown in Table~\ref{tab:result_example}. 

In Example~(a) and (b), Both models succeeded to recognize the entity span.
However, only the proposed model also correctly predicted the type label.
Note that the entities \texttt{Location/Spa} and \texttt{Natural\_Object/Living\_Thing/Living
\_Thing\_Other} 
appear rarely, but rather to the extent of the top layer components \texttt{Location} and \texttt{Natural\_Object} that appear frequently in the training set.
Therefore, these examples suggest that the proposed model effectively exploits shared information of label components, especially in terms of the hierarchical layer.

Although, we found that the proposed model predicts partially correct labels even though it is not totally correct in some cases.
In Example~(c), あお白い (\textit{pale}) is categorized into \texttt{Color/Color\_Other}, the proposed model also predicted the wrong label \texttt{Color/Nature\_Color}.
However, interestingly, the proposed model correctly recognized the top layer of the type label as \texttt{Color}, which is in contrast to the completely wrong prediction of the baseline model.
\end{document}